\pdfoutput=1
\documentclass[11pt]{article}

\usepackage{EMNLP2022}

\usepackage{times}
\usepackage{latexsym}

\usepackage[T1]{fontenc}
\usepackage[utf8]{inputenc}

\usepackage{microtype}

\usepackage{inconsolata}

\usepackage{color}
\usepackage{url}
\usepackage{multirow}
\usepackage{graphicx}

\title{Dealing with Abbreviations in the Slovenian Biographical Lexicon}

\author{Angel Daza$^1$\quad Antske Fokkens$^1$\quad Toma\v{z} Erjavec$^2$\quad \medskip\\
  $^1$CLTL, Vrije Universiteit Amsterdam\\
  $^2$Department of Knowledge Technologies, Jožef Stefan Institute \\
  \medskip
  \texttt{\{j.a.dazaarevalo,antske.fokkens\}@vu.nl, tomaz.erjavec@ijs.si}}

\begin{document}
\maketitle
\begin{abstract}
Abbreviations present a significant challenge for NLP systems because they cause tokenization and out-of-vocabulary errors. They can also make the text less readable, especially in reference printed books, where they are extensively used. Abbreviations are especially problematic in low-resource settings, where systems are less robust to begin with. In this paper, we propose a new method for addressing the problems caused by a high density of domain-specific abbreviations in a text. We apply this method to the case of a Slovenian biographical lexicon and evaluate it on a newly developed gold-standard dataset of 51 Slovenian biographies. Our abbreviation identification method performs significantly better than commonly used ad-hoc solutions, especially at identifying unseen abbreviations. We also propose and present the results of a method for expanding the identified abbreviations in context. % and show that its application substantially improves results of an out-of-the-box NER-system on our data.
\end{abstract}

\section{Introduction}

Abbreviations such as "b." for "born", or "gr." for "graduated" are a common issue when dealing with digitized texts which use a large number of them for space-saving reasons.
They are also a known problem when processing technical documents \cite{park-byrd-2001-hybrid} and biomedical texts \cite{jin-etal-2019-deep-bio-abbr}. 
% and parliamentary transcriptions \cite{zelasko-2018-expanding}.
In this paper, we examine the case of biographical dictionaries, i.e.\ collections of biographies that have been digitized, and, in particular, the Slovenian Biographical Lexicon. 

To automatically extract facts from biographical texts, Digital Humanities researchers normally rely on out-of-the-box NLP tools such as Stanza \cite{qi-etal-2020-stanza} or SpaCy.\footnote{\url{https://spacy.io/}} These tools are often adequate for identifying sentences which are then used as input for higher-level downstream tasks, for manual inspection, and for visualization purposes. However, out-of-the-box tools are designed to work for the broadest possible text domains and cover the most common cases. This impacts performance significantly when dealing with domain-specific data, such as entries of biographical dictionaries, and more so when they contain a lot of abbreviations. The problem is even more pronounced when dealing with a relatively lower resource language, such as Slovenian. The performance bottleneck occurs already at the first step, i.e. tokenization: in order to perform good tokenization in domain-specific texts, we need to have a reliable method for identifying abbreviations such that the tokenizer does not split them wrongly, generating faulty tokens and incomplete sentences. In this paper we:

\begin{itemize}
    \setlength\itemsep{-0.5em}
    \item Quantify the effect that abbreviations have on a downstream task such as NER on Slovenian biographical texts. 
    \item Propose a method for abbreviation identification, apply it to raw texts and compare it to straightforward baselines.
    \item Analyze the feasibility of using contextually dependent word embeddings, in particular, the SloBERTa \cite{Ulcar_Sloberta:21} language model, to automatically expand abbreviations in text and improve readability.
    \item Evaluate the performance of our methods on a new human-curated dataset with, inter alia, gold tokens, sentences, named entities, and expanded abbreviations.\footnote{\url{https://github.com/angel-daza/abbreviation-detector}}
\end{itemize}

\section{Related Work}\label{related_work}

Specific work addressing abbreviations is scarce. We think this is due to the fact that it is addressed as a preprocessing step with tailored solutions for each specific use case, involving regular expressions, or corpus-specific rules \cite{bollmann-etal-2011-rule}. A few papers try to construct methods for a general solution to this problem. For instance, \newcite{park-byrd-2001-hybrid} propose a pipeline system to induce acronyms, where every sequence of characters (separated by spaces) is considered a \textit{candidate abbreviation} if it satisfies certain conditions. \citet{zelasko-2018-expanding} proposes a more advanced approach based on an LSTM classifier that uses morphosyntactic information about a sentence to directly infer the correct expansion of an abbreviation in Polish. Direct work on abbreviations also exists in the biomedical domain, including detection and disambiguation \cite{stevenson-etal-2009-bio-abbr} as well as abbreviation expansion \cite{jin-etal-2019-deep-bio-abbr}. Finally, \citet{gorman-etal-2021-structured-abbreviation} have recently developed an English dataset to explore abbreviation expansion methods taking into account the context.

Another common way to address the difficulty of domain-specific texts is \textit{\textit{text normalization}}, which is the task of \textit{translating} a domain-specific text into more standard form (this can be at different levels such as lexical or morphological) that is easier to process by general purpose NLP tools. This approach is common when dealing with user generated text and social media \cite{pennell-liu-2011-character, baldwin-etal-2015-shared, van-der-goot-2019-monoise}, and also historical texts \cite{scherrer-erjavec-2013-modernizing, Ljubesic2016NormalisingSD, bollmann-2019-large}.

One drawback of text normalization is that it is frequently implemented using an Encoder-Decoder approach \cite{robertson-goldwater-2018-evaluating,bollmann-etal-2019-shot}, which requires a big-enough parallel corpus to be trained and obtain good-quality results. Another drawback is the fact that it \textit{generates} a new standard sentence, which is not a desired side effect if we want to preserve word by word the original biographical text. In contrast to the normalization task, we are interested in preserving the original text and only identifying (and perhaps expanding) the abbreviations that are problematic for the NLP tools.

\section{Dataset}\label{dataset}

%ET: it would be nice to put this in, if space allows:
%For space saving reasons the biographies contain many abbreviations, as illustrated in the following snippet, where %every fifth word is an abbreviation:
%
%\indent \textit{Agustich (Augustič) Imre, prekmurski pisatelj, r. 29. sept. 1837 v Murskih Petrovcih v Železni županiji, 
%u. v Budimpešti 17. jul. 1879. Šest gimn. razr. je dovršil v Sobotišču ...}

The Slovenian Biographical Lexicon (SBL) was published in 15 volumes (1925--1991) and contains 5,047 biographies \cite{sbl}. 
% It is fully digitised, encoded as a TEI document, and is available on the Web in the scope of the Slovenian Biography portal \cite{sbl}. 
For the experiments described in this paper, we have created the dataset SBL-51abbr \cite{11356/1588}, which consists of 51 randomly selected entries from SBL\footnote{\url{http://hdl.handle.net/11356/1588}}.
The text of each entry is manually tokenized and sentence segmented, marked with named entities, and lemmatized words. It has also been automatically annotated with Universal Dependencies PoS tags, morphological features and dependency parses using CLASSLA~\cite{ljubesic-dobrovoljc-2019-neural},\footnote{\url{https://pypi.org/project/classla/}} a fork of the Stanford Stanza pipeline \cite{qi-etal-2020-stanza},\footnote{\url{https://stanfordnlp.github.io/stanza/}} which is the state-of-the-art tool for annotating Slovenian. Crucially for the envisaged use of the corpus, the abbreviations in the corpus have been manually expanded so that the expansions are in the correct inflected form.
% given their sentence context.
The curated dataset consists of 655 sentences (see Table~\ref{tab:dataset_table}). It is available in the canonical TEI encoding, and derived plain text and CoNLL-U files. The plain-text file has abbreviations and their expansions marked up with [[...]]((...)) respectively. There are two CoNLL-U files, one with the text stream with abbreviations, and one with the text stream with expansions. Note that only the one with expansions has syntactic parses. Both CoNLL-U files have the expansions / abbreviations and named entities marked up in IOB format in the last column.

% and is available in several formats including the CoNLL-U format. 
% Information about the named entities and the abbreviations is given in the IOB format in the MISC (last) column. 
We use this dataset as a gold standard to test the performance of our proposed methods. We randomly split the available data into three portions: 70\% for training, 10\% for development and 20\% for testing. 

\begin{table}[htb]
\centering
\begin{tabular}{lrrrr}
Split              & Sents      & Abbrs   &  Unique  & Unseen \\\hline
Train              & 458        & 1385    &  399   & 0 \\
Dev                &  66        &  236    &  130   & 33 \\
Test               & 131        &  420    &  181   & 70 \\ \hline
$\Sigma$           & 655        & 2041    &  710   &
\end{tabular}
\caption{Abbreviation statistics on the SBL-51abbr dataset. We count the total number of abbreviations, the unique types and the number of unseen abbreviations in the dev and test splits.}
\label{tab:dataset_table}
\end{table}
\section{Impact of Abbreviations}\label{abbrev_impact}

We first quantify the impact that the high number of abbreviations in the SBL-51abbr corpus has when processing the raw texts with CLASSLA~\cite{ljubesic-dobrovoljc-2019-neural} and performing NER. We compare the performance on the original texts (with all abbreviations) with a second scenario where abbreviations were substituted with their gold expansions. Table~\ref{tab:NERvsNER} shows the performance per class when processing the original version (the first row of numbers per label), and right below is the performance when processing the same texts but without any abbreviation. There is a significant boost all across the board for the fully expanded texts, resulting on over 30 F1 points of improvement on the macro average measure. This shows that having effective methods for abbreviation identification and expansion can be beneficial.

\begin{table}[htb]
\centering
\begin{tabular}{ccccc}
\textbf{Label}              & \textbf{P} & \textbf{R} & \textbf{F1}    \\ \hline
\multirow{2}{*}{PER}        & 68.75      & 22.45      & 33.85        \\
                            & 76.80       & 65.31      & 70.59                    \\ \hline
\multirow{2}{*}{DERIV-PER}  & 50.00     & 4.76       & 8.70  \\
                            & 92.86      & 61.90       & 74.29             \\ \hline
\multirow{2}{*}{LOC}        & 85.29      & 39.73      & 54.21 \\
                            & 82.32      & 92.47      & 87.10    \\ \hline
\multirow{2}{*}{MISC}       & 65.38      & 12.41      & 20.86   \\
                            & 56.14      & 23.36      & 32.99          \\ \hline
\multirow{2}{*}{ORG}        & 23.53      & 14.81      & 18.18   \\
                            & 22.06      & 55.56      & 31.58            \\ \hline
\multirow{2}{*}{macro\_avg} & 58.59      & 18.83      & 27.16  \\
                            & 66.03      & 59.72      & 59.31     \\  \hline              
\end{tabular}
\caption{NER scores when applied to the original sentences with abbreviations (upper rows) vs sentences with all abbreviations expanded (lower rows).}
\label{tab:NERvsNER}
\end{table}
\section{Dealing with Abbreviations}\label{abbreviation_identif_expan}

\begin{figure*}
\centering
  \includegraphics[width=0.85\textwidth]{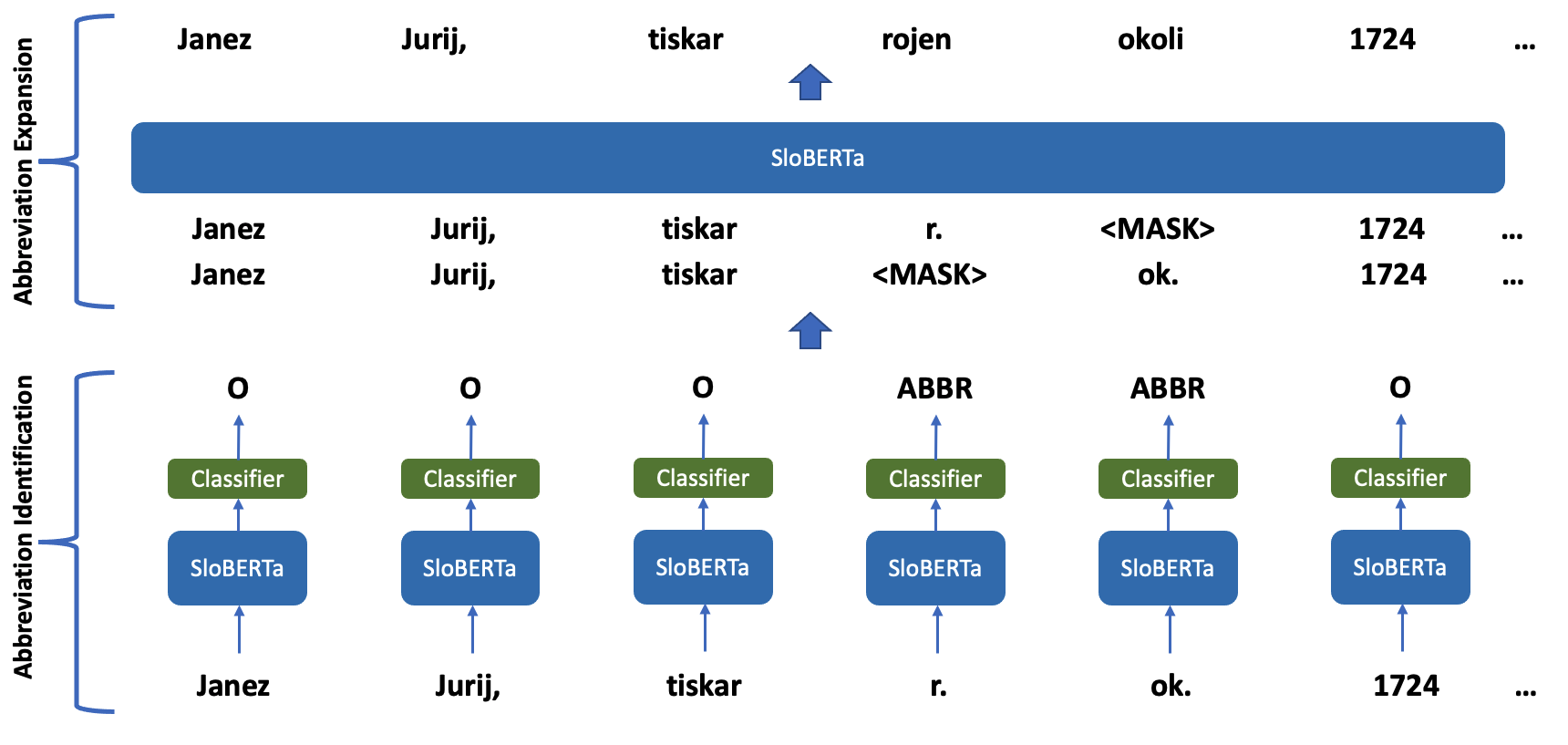}
  \caption{We first train a classifier for identifying abbreviations (SloBERTa+Neural linear layer) and then match each of the identified candidates to use SloBERTa as a predictor for that slot. We only keep the suggested expansion if it meets the requirements.}
\label{fig:method}
\end{figure*}

\subsection{Baselines}

\textbf{Dictionary-based:} to bypass tokenizer-specific noise, we split every document by spaces to obtain a list of \textit{dirty tokens}.\footnote{We call them \textit{dirty} because they will have punctuation attached to them. For example "Hello world!" will be \textit{tokenized} as ['Hello', 'world!'] instead of ['Hello', 'world', '!'] which would be the optimal tokenization.} For each token, we first clean it, meaning we remove all special characters except full stops. If the \textit{clean token} does not end with a full stop we skip it, otherwise we strip the full stop and check if the entry exists in a large dictionary. We use two dictionaries: the Hunspell dictionary,\footnote{\url{https://github.com/hunspell/hunspell}} a popular tool used for spelling correction, and GigaFida 2.0 \cite{krek-etal-2020-gigafida}, a big reference corpus of standard Slovene. Because dictionaries only contain entries for complete words, we consider the token an abbreviation if no entry exists.

\textbf{Corpus-based:} we tokenize the training corpus using CLASSLA and compute the frequency for all token unigrams $t_1$ and bigrams $(t_1,t_2)$ in the corpus. If a bigram contains a full stop as a second component or if a unigram has a full stop as its last character,\footnote{We test for this case because, if the tokenizer rightly recognized the abbreviation, then the full stop will still be attached to it} then we increase the count of $t_1$ in $\mathcal{A}$ otherwise we increase the count of $t_1$ in $\mathcal{B}$. We take all $t_1$s that appear in both lists and calculate their probability to be an abbreviation as:

\begin{equation}
    P(t_1=abbr)=\frac{freq(t_1 \epsilon \mathcal{A})}{freq(t_1 \epsilon \mathcal{A})+freq(t_1 \epsilon \mathcal{B})}
\end{equation}

If the probability $P(t_1=abbr) \geq 0.8 $ then $t_1$ is considered to be an abbreviation, otherwise we skip it. This method will, of course, carry over some of the tokenization mistakes. The reasoning behind this baseline is to capture the number of times a given token appears before a full stop compared to the total times it appears in the corpus. If it is the case that most of the occurrences of such a token are immediately followed by a full stop, then it is most likely an (unrecognized) abbreviation.

\subsection{Abbreviation Classifier}
We propose an automatic method for identifying abbreviations by fine-tuning a classifier on top of the SLoBERTa language model \cite{Ulcar_Sloberta:21}. We again obtain the sequence of tokens by splitting the raw texts by spaces. We treat each one of the \textit{dirty tokens} as a separate input sequence to SloBERTa. We train the classifier using the gold abbreviation labels to predict if a token is an abbreviation or not (Figure \ref{fig:method}, bottom).

% \begin{figure}
% \centering
%   \includegraphics[width=0.5\textwidth]{figs/SloBERTa2.png}
%   \caption{We first train a classifier for identifying abbreviations (SloBERTa+Neural linear layer) and then mach each of the identified candidates to use SloBERTa as a predictor for that slot. We only keep the suggested expansion if it meets the requirements.}
% \label{fig:method}
% \end{figure}

\subsection{Abbreviation Expansion}
Once we have our text with abbreviation candidates identified, for each candidate, we take the full sentence it appears in, mask it and let SloBERTa predict the masked token. We take SloBERTa's prediction to be a valid expansion if one of the top~5 predicted candidates starts with the same letter as the masked abbreviation, otherwise we leave the original abbreviation. This way we are substituting each candidate \textit{in-context} and thus approaching the optimal scenario with the fully expanded sentences where NER performed much better (see Table~\ref{tab:NERvsNER}). A visualization of both the identification and expansion steps of our method is given in Figure~\ref{fig:method}.
\section{Results and Evaluation}\label{evaluation}

\subsection{Abbreviation Identification Baselines}

We first present the results on the test set obtained by our proposed baselines in Table \ref{table:baselines}. These baselines represent common pre-processing approaches to dealing with abbreviations. We can see that the two dictionary versions suffer from low recall, especially the GigaFida dictionary with only 20\%. This behavior is expected since we are dealing with a domain-specific (and partially historical) text. The second baseline behaves much better and achieves an 85.34 F1 score. The Bigrams+Dict version mixes both approaches, which improves the coverage of identified abbreviations, but unfortunately lowers the high precision of the bigram approach.

\begin{table}[htb]
\centering
\begin{tabular}{lllll}
\textbf{Baseline}                & \textbf{P}     & \textbf{R}    &  \textbf{F1} \\ \hline
GigaFida Dict	        & 89.36	        & 20.00	    & 32.68 \\
Hunspell Dict	        & 80.81	        & 71.19	    & 75.70 \\
Corpus Bigrams	        & 95.85	        & 76.90	    & 85.34 \\
Bigrams+Dict	        & 73.27	        & 95.95	    & 83.09
\end{tabular}
\caption{Abbreviation identification baseline results on the test set. They show trade-offs between precision and recall. The bigrams method performs the best.}
\label{table:baselines}
\end{table}

\subsection{Abbreviation Classifier} 

The baselines show a trade-off between good precision or good recall. In contrast, Table \ref{table:sloberta} shows that our SloBERTa method significantly increases the recall without hampering precision. We fine-tuned SloBERTa\footnote{We used the default settings from HuggingFace \url{https://huggingface.co/EMBEDDIA/sloberta}} 
for 5 epochs and pick the model that performs best on the development set. We present the mean of 5 experiments with different random seeds together with the standard deviation. The results demonstrate that this is a stable approach for identifying abbreviations in text.

\begin{table}[htb]
\centering
\begin{tabular}{cccc}
\textbf{Split }  &  \textbf{P}     &    \textbf{R}            &   \textbf{F1}               \\\hline
Dev     &   95.91$^{\pm 0.75}$ 	   &    97.91$^{\pm 1.9}$ 	  &   96.89$^{\pm 0.9}$        \\
Test    &   $93.94^{\pm 1.5}$ 	   &    98.10$^{\pm 2.0}$     &   95.97$^{\pm 1.3}$        \\
\end{tabular}
\caption{Abbreviation identification results with our SloBERTa binary classifier. Results on test are 10 points above the best baseline.}\label{table:sloberta}
\end{table}

\subsection{Abbreviation Expansion} 
We measure the success of our abbreviation expansion method by re-running the NER tagger on the sentences with the expanded abbreviations as predicted by SloBERTa (see Table \ref{tab:NER_expanded}). From the 420 abbreviations in the test set, 154 where expanded following our heuristic and the rest of abbreviations in the sentences were left untouched. We can compare these scores directly with our analysis from Table~\ref{tab:NERvsNER} and see that even though our method for expanding is quite basic, it already gets us closer to the ceiling scores (where all gold expansions were substituted).  Important gains can be seen in all categories and the macro average score reached with the predicted expansions is 49.64 F1 which is 22 points above the 27.16 F1 obtained originally.

\begin{table}[htb]
\centering
\begin{tabular}{cccc}
\textbf{Label} & \textbf{P} & \textbf{R} & \textbf{F1} \\ \hline
PER            & 40.54      & 67.67      & 50.70       \\
DERIV-PER      & 78.57      & 55.00      & 64.71       \\
LOC            & 72.33      & 82.73      & 77.18       \\
MISC           & 34.34      & 27.64      & 30.63       \\
ORG            & 17.14      & 46.15      & 25.00       \\\hline
macro\_avg     & 48.59      & 55.84      & 49.64      
\end{tabular}
\caption{The NER results on the test set after applying SloBERTa-based expansions show consistent improvements compared to the original sentences (cf. Table \ref{tab:NERvsNER})}
\label{tab:NER_expanded}
\end{table}
\section{Conclusions}\label{conclusions}

In this paper we focused on the task of Named Entity Recognition to quantify the impact of abbreviations in a text by comparing the performance of the CLASSLA NER tagger on the same sentences with and without abbreviations. We presented a gold-standard dataset consisting of 51 biographies in Slovenian (a limited-resource language) in a specialized text domain. We also presented a method for automatically identifying abbreviations and expanding them without the need for a tokenizer. The biggest advantage of our method is that it can be applied out-of-the-box for any language which has a large language model available and does not need ad-hoc training data or large fixed dictionaries. Our abbreviation identification classifier obtains better precision and better recall when compared to other straightforward approaches to identify abbreviations. 

Finally, we presented a method that uses a pre-trained language model to predict plausible expansions for the identified abbreviations. We notice that our method is still simple but already achieves better results than directly processing the original sentences with abbreviations. In future work we aim to explore more sophisticated methods for abbreviation expansion that allow us to further improve the readability of texts. We find our results encouraging for researchers working with limited domains who may find a similar approach helpful for improving performance in other tasks.

% Even though our expansion method is far from ideal, the fact that the abbreviation classifier has such good results, allows us to e.g. add the abbreviations as exceptions to the tokenizers.

\section*{Limitations}

The results presented in this paper have been evaluated on the specific use case of the Slovenian Biographical Lexicon. When considering to apply this approach to other use cases, the following limitations should be taken into account:
\paragraph{Good Use Case.} The SBL is a good use case in the sense that this is a domain with a high density of abbreviations and thus a relatively high number of positive class examples. This means that the relatively small dataset was comparatively rich (i.e.\ another domain may require more data) and the potential of improving results is relatively high (i.e.\ identifying abbreviations may have less impact on downstream tasks in other domains). 
% Moreover, despite its relatively low resource status, there is a reasonable performing NER system and contextualized language model available for Slovenian. 
\paragraph{Language Model Required.} A large language model is needed for this approach and may not be available for many low or even medium resource languages. This is unfortunate, because this research aims to support relatively low resource languages that must rely on standard tools, because there are limited resources for creating new data sets and models.
\paragraph{Expanding Abbreviations Remains Largely Unsolved.} The results for expanding abbreviations are still meagre. Even though the current approach is simple, it may already represent an upperbound due to the productive character of abbreviations in this domain, the rich inflection of Slovenian, and the considerable effort required to obtain more training data.

% \section*{Limitations}
% EMNLP 2022 requires all submissions to have a section titled ``Limitations'', for discussing the limitations of the paper as a complement to the discussion of strengths in the main text. This section should occur after the conclusion, but before the references. It will not count towards the page limit.  

% The discussion of limitations is mandatory. Papers without a limitation section will be desk-rejected without review.
% ARR-reviewed papers that did not include ``Limitations'' section in their prior submission, should submit a PDF with such a section together with their EMNLP 2022 submission.

% While we are open to different types of limitations, just mentioning that a set of results have been shown for English only probably does not reflect what we expect. 
% Mentioning that the method works mostly for languages with limited morphology, like English, is a much better alternative.
% In addition, limitations such as low scalability to long text, the requirement of large GPU resources, or other things that inspire crucial further investigation are welcome.

% \section*{Ethics Statement}
% Scientific work published at EMNLP 2022 must comply with the \href{https://www.aclweb.org/portal/content/acl-code-ethics}{ACL Ethics Policy}. We encourage all authors to include an explicit ethics statement on the broader impact of the work, or other ethical considerations after the conclusion but before the references. The ethics statement will not count toward the page limit (8 pages for long, 4 pages for short papers).

\section*{Acknowledgements}
This work was partially supported by the EU Horizon 2020 project InTaVia: In/Tangible European Heritage - Visual Analysis, Curation and Communication (\url{http://intavia.eu}) under grant agreement No. 101004825.

\bibliography{emnlp2022}
\bibliographystyle{acl_natbib}

% \appendix
% \section{Example Appendix}\label{sec:appendix}
% This is a section in the appendix.

\end{document}